\documentclass[conference]{IEEEtran}
\IEEEoverridecommandlockouts
\usepackage{cite}
\usepackage{amsmath,amssymb,amsfonts}
\usepackage{algorithmic}
\usepackage{graphicx}
\usepackage{textcomp}
\usepackage{xcolor}
\usepackage{fancyvrb}
\def\BibTeX{{\rm B\kern-.05em{\sc i\kern-.025em b}\kern-.08em

    T\kern-.1667em\lower.7ex\hbox{E}\kern-.125emX}}
\begin{document}

\title{Statistical Properties of the log-cosh Loss Function Used in
Machine Learning\\

}

\author{\IEEEauthorblockN{ Resve A. Saleh}
\IEEEauthorblockA{\textit{Dept. of Electrical and Computer Engineering} \\
\textit{University of British Columbia}\\
Vancouver, Canada \\
res@ece.ubc.ca}
\and
\IEEEauthorblockN{ A.K.Md. Ehsanes Saleh}
\IEEEauthorblockA{\textit{School of Mathematics and Statistics} \\
\textit{Carleton University}\\
Ottawa, Canada \\
esaleh@math.carleton.ca}
}

\maketitle

\begin{abstract}
This paper analyzes a popular loss function used in machine learning called the log-cosh loss function. A number of papers have been published using this loss function but, to date, no statistical analysis has been presented in the literature.
In this paper, we present the distribution function from which the log-cosh loss arises. We compare it to a similar distribution, called the Cauchy distribution,
and carry out
various statistical procedures that characterize its properties.
In particular, we examine its associated pdf, cdf, likelihood function
and Fisher information.
Side-by-side we consider the Cauchy and Cosh distributions as well as the
MLE of the location parameter with
asymptotic bias, asymptotic variance, and confidence intervals.
We also provide a comparison of robust estimators from several other loss functions,
including the Huber loss function and the rank dispersion function.
Further, we examine the use of the log-cosh function for quantile regression. 
In particular,
we identify a quantile distribution function from which
a maximum likelihood estimator for quantile regression can be derived.
Finally, we compare a quantile M-estimator based on log-cosh with robust monotonicity
against
another approach to quantile regression based on convolutional smoothing.
\end{abstract}


\section{Introduction}
According to several authors \cite{b1}\cite{b2}\cite{b12}\cite{b21}, the log-cosh loss function is one of the 
most important loss functions in machine learning today
but very little has been published about its statistical
characteristics. It has been implemented in many software
environments for machine learning and used in a number of different research
efforts.
From a programming perspective, it may be found in R in the \texttt{limma} package \cite{b3} and
can be implemented
in a single line of python using the \texttt{numpy} library \cite{b7}.
It is also available in TensorFlow2 in the Keras library \cite{b4}, 
and likewise in PyTorch \cite{b15} for deep learning.

Research areas of application include 
variational autoencoders \cite{b11}\cite{b8} and cancer detection \cite{b17}.
It has also been used in
tree-based learning algorithms such as XGBoost\cite{b12}, and most recently for a 
solution to the crossing problem in quantile regression \cite{b5}.
One caveat is that a function like $\log(\cosh(x))$ can overflow if
not implemented correctly. However, this issue has been
addressed through built-in library functions
and the problem is rarely encountered. As a result, it has been routinely used
in machine learning for a number of years now.

The log-cosh loss function belongs to the class of robust estimators
that tend to prefer solutions in the vicinity of the median
rather than the mean. Another view is that robust estimators
are more tolerant to outliers 
in the data set and this is perhaps one
of the key reasons to select the log-cosh loss function over others. It is also
continously differentiable, unlike the Huber loss function which does
not have a continuous second derivative. These aspects are well-known in
the machine learning community. The
gap is the literature on log-cosh is relative to its origin and
statistical properties.
In this paper, we seek to remedy this gap by deriving the log-cosh loss function
from first principles,
starting with its distribution, and studying properties such as bias, variance,
confidence intervals, Fisher information and standard errors of estimation.
\section{M-estimators}
The log-cosh loss function falls in the category of M-estimators having the
general form:
\begin{equation}
Q(x,\theta)=\frac1n \sum_{i=1}^n \rho(x_i,\theta)
\end{equation}
where $\rho(x_i,\theta)$ is a term in a given loss function. Here $X$ is an iid random
variable associated with the residuals and $\theta$ is the parameter to be estimated.
Any desired function may be used
for $\rho(x_i,\theta)$ as long as it satisfies some minimal set of conditions.
The estimate of $\theta$ is produced as follows:
\begin{equation}
\hat\theta = \text{argmin}_{\theta\in \mathbb{R}} \frac1n \sum_{i=1}^n \rho(x_i,\theta).
\end{equation}
This is typically carried out using numerical optimization
employing some form of gradient descent in the context of machine learning.
Of particular interest here is the log-cosh loss function given by
\begin{equation}
\rho_L(x_i,\theta) = \text{log}(\text{cosh}(x_i-\theta)).
\end{equation}

In the sections to follow, we show how to derive this loss function
starting with a probability distribution and then provide a more
intuitive view of its origin.

\section{Background}
\subsection{Hyperbolic Functions}
Recall the basic hyperbolic functions as follows:

$$\text{sinh}(x) = \frac{e^x-e^{-x}}{2}$$

$$\text{cosh}(x) = \frac{e^x+e^{-x}}{2}$$
and
$$\text{tanh}(x) = \frac{e^x-e^{-x}}{e^x+e^{-x}}.$$

We also have that
$$\text{sech}(x) = \frac1{\text{sinh}(x)}=\frac{2}{e^x-e^{-x}}$$

$$\text{csch}(x) =\frac1{\text{cosh}(x)}= \frac{2}{e^x+e^{-x}}$$
and
$$\text{coth}(x) =\frac1{\text{tanh}(x)}= \frac{e^x+e^{-x}}{e^x-e^{-x}}.$$
\subsection{The Cosh Distribution}
\begin{figure}[!ht]
    \centering
    \includegraphics[scale = 0.4]{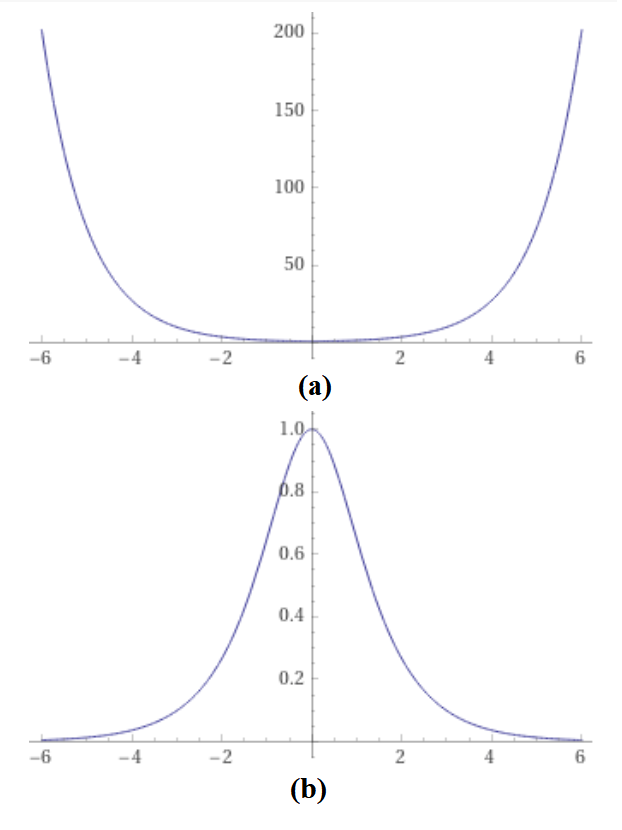}
    \caption{(a) $\text{cosh}(x)$  (b) $1/\text{cosh}(x)$}\label{fig:cosh}
\end{figure}

Of primary interest here is the cosh(x) function
which is plotted in Fig. \ref{fig:cosh}(a).
The reciprocal of this function, i.e., $\text{csch}(x)$, is plotted in Fig. \ref{fig:cosh}(b). 
This function resembles a probability density function (pdf) but in order to
satisfy the necessary conditions, it must be non-negative and integrate to 1. 
The first requirement is already satisfied by inspection. For the
second requirement, if we take the
integral of the function, we obtain the normalizing constant as

$$\int_{-\infty}^{\infty} \frac{1}{\cosh(x)} dx = \pi.$$

Therefore, the pdf of the Cosh distribution is given by
\begin{equation}\label{eqn21}
f(x) =  \frac{1}{\pi\,\cosh(x)}
\end{equation}

\begin{figure}[!ht]
    \centering
    \includegraphics[scale = 0.4]{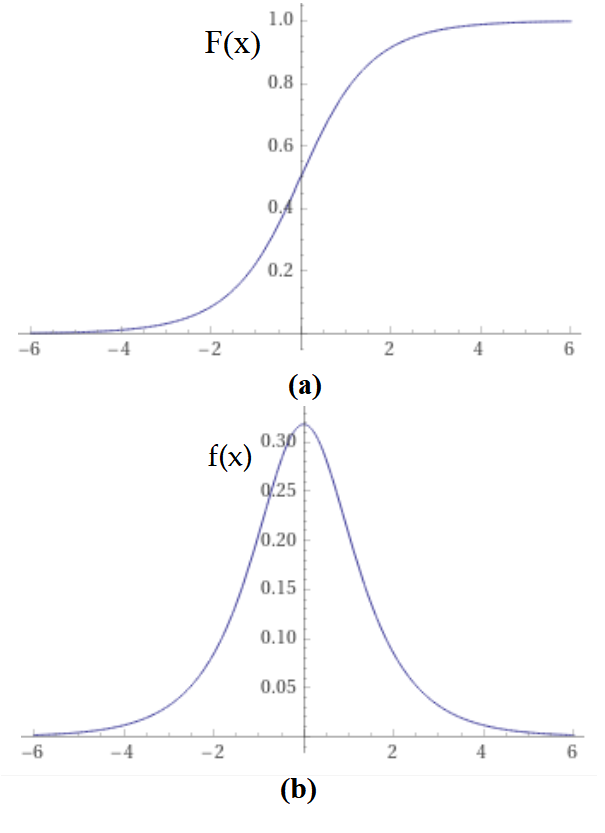}
    \caption{Cosh distribution (a) cdf  (b) pdf}\label{fig:pdfcdf}
\end{figure}

The corresponding cumulative distribution function (cdf) is given by:
\begin{equation}\label{eqn01}
F(x) =  \frac12+ \frac1\pi {\text{tan}^{-1}\big[\sinh\big({x}\big)}\big]
\end{equation}
which is invertible.
The cdf and pdf are plotted in Fig. \ref{fig:pdfcdf}.

More generally, we have a location-scale family of distributions of the form
\begin{equation}\label{eqn1}
f(x;\theta,\sigma) =  \frac{1}{\pi\sigma\,\text{cosh}(\frac{x-\theta}{\sigma})}
\end{equation}
where $\theta$ is the location parameter and $\sigma$ is the scale parameter.
We note here that
$$\mathbb{E}[X] = \theta\,\,\,\,\,\text{and}\,\,\,\,\, \mathbb{E}[X^2]=\frac{\pi^2\sigma^2}{4}+ \theta^2 $$
so the asymptotic variance is
$$\text{Var}(X) = \mathbb{E}[X^2]-\mathbb{E}[X]^2 =\frac{\pi^2\sigma^2}{4}.$$
The cdf given by:
\begin{equation}\label{eqn01}
F(x;\theta,\sigma) =  \frac12+ \frac1\pi \tan^{-1}\bigg[\sinh\bigg(\frac{x-\theta}{\sigma}\bigg)\bigg].
\end{equation}
\subsection{Maximum Likelihood Estimation}
We next develop the maximum likelihood estimator (MLE) for the case of $\sigma=1$,
as follows. If $x_1, x_2,...,x_n$ are i.i.d random variables, then the likelihood function is given by:
\begin{equation}\label{eqn3}
L(x_1, x_2,...,x_n;\theta) = \prod_{i=1}^n \frac{1}{\pi\,\text{cosh}(x_i-\theta)}.
\end{equation}
The negative log-likelihood expression is given by:
\begin{equation}\label{eqn4}
-\ell (x_1, x_2,...,x_n;\theta) =  \text{log}\pi +\sum_{i=1}^n\text{log}(\text{cosh}(x_i-\theta)).
\end{equation}
The estimator $\hat\theta$ is the solution to the equation:
\begin{equation}\label{eqn5}
\sum_{i=1}^n \tanh(x_i-\theta) = 0
\end{equation}
or equivalently
\begin{equation}\label{eqn060}
\hat \theta =  \text{argmin}_{\theta \in \mathbb{R}} \sum_{i=1}^n\text{log(cosh}(x_i-\theta)).
\end{equation}
We see that this estimator is, in fact, the MLE of the Cosh distribution.
The equation can be solved in a straight-forward manner
using a convex optimization procedure due to 
the fact that it is globally convex. That is, the second
derivative can be shown to be non-negative.
This can be demonstrated by noting that
\begin{equation}
\psi_L(x) = \rho_L'(x) = \tanh(x)
\end{equation}
and 
\begin{equation}
\psi_L'(x)=\rho_L''(x) = \text{sech}^2(x).
\end{equation}
Therefore, both the gradient and Hessian terms can be easily obtained
and this is important for machine learning applications.
\section{An Intuitive View of log-cosh}\label{AA}
It is useful to consider the log-cosh loss function from
an intuitive standpoint to understand its relative importance.
We first examine the loss function in terms of its behavior
relative to the L1 and L2 functions.
Consider M-estimators of the form
\begin{equation}
\hat\theta = \text{argmin}_{\theta\in \mathbb{R}} \frac1n \sum_{i=1}^n \rho(x_i,\theta).
\end{equation}
For example, we can define $\rho(x)$ in any manner we choose such as
\begin{equation}
\rho_{L2}(x) = x^2
\end{equation}
for least squares estimate (LSE), also called the L2 loss function, and
\begin{equation}
\rho_{L1}(x,\theta) = |x|
\end{equation}
for least absolute deviation (LAD), also called the L1 loss function. 
For log-cosh, we already derived the MLE to be
\begin{equation}
\rho_L(x) = \text{log}(\text{cosh}(x))
\end{equation}
Next, we compare the log-cosh loss function relative to the L1 and L2 cases above.
As $x \rightarrow +\infty$, the log-cosh loss tends towards
\begin{equation}\label{eqn061}
\text{log(cosh}(x)) \approx x - \text{log}(2)
\end{equation}
and as $x \rightarrow -\infty$, the function tends towards 
\begin{equation}\label{eqn062}
\text{log(cosh}(x)) \approx - x - \text{log}(2).
\end{equation}
Therefore, far from $x=0$, the function behaves as
\begin{equation}\label{eqn063}
\text{log(cosh}(x)) \approx |x| - \text{log}(2)
\end{equation}
which is like L1 except for the constant term.
Now, as $x \rightarrow 0$, we can take a Taylor series expansion at $x=0$ as follows
\begin{equation}\label{eqn064}
\text{log(cosh}(x)) = \frac{x^2}{2}-\frac{x^4}{12}+\frac{x^6}{45}+O(x^8)
\end{equation}
and therefore it is, to first-order, like L2. That is,
\begin{equation}\label{eqn065}
\text{log(cosh}(x)) \approx \frac{x^2}{2}
\end{equation}
for small $x$.
Since it behaves like L2 close to the origin and L1 far from the origin,
one can view it as a smoothed out L1 using L2 around the origin. Hence,
both the Jacobian and Hessian matrix terms exist for this loss function.
For robust regression, this function can be used as an alternative to L1.

Huber \cite{b26} proposed a method of combining the best of L1 and L2 by explicitly
using L2
in the vicinity of the origin where the discontinuity lies, and then 
switching to L1 a certain distance, $\delta$, away from the origin.
The small interval around the
origin is defined by $[-\delta,+\delta]$. Inside this region, the L2 loss
function is used since it is continuous. Outside this region, the L1 loss
is used but great care is taken to match the derivatives 
at the interface between the two
regions. The resulting Huber loss function is given by:
\begin{equation}\label{eqn:huber0}
\hat\theta_n^{\rm H} = \text{argmin}_{\theta} \sum_{i=1}^n\rho_H(x_i - \theta)
\end{equation}
where
\begin{equation}\label{eq:ch1:huber1}
\rho_H(x) =
  \begin{cases}
    x^2/2       & \quad \text{if } |x| \le \delta\\
    \delta(|x|-\frac{\delta}{2})  & \quad \text{if } |x| > \delta
  \end{cases}
\end{equation}
The derivative of the Huber function is given by:
\begin{equation}\label{eq:ch1:huber2}
 \psi_H(x) =
  \begin{cases}
    x       & \quad \text{if } |x| \le \delta\\
    +\delta       & \quad \text{if } x > +\delta\\
    -\delta & \quad \text{if } x < -\delta
  \end{cases}
\end{equation}

\begin{figure}[!ht]
  \centering
  \includegraphics[scale = 0.9]{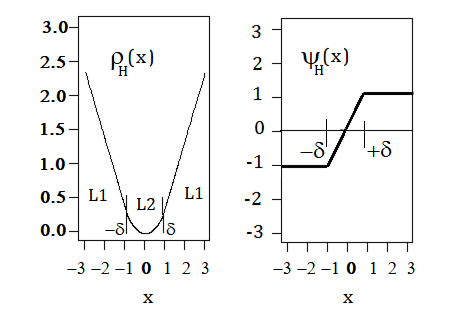}
  \caption{Huber loss and derivative as a function of $x$ for $\delta=1$. }\label{fig:ch1:4:Huberplots}
\end{figure}

There are a total of 3 regions defined by the Huber function:  $x < - \delta$, $x > + \delta$, and $|x|<\delta$. As shown in Figure \ref{fig:ch1:4:Huberplots}, the Huber function and its first derivative are both continuous
in all regions. In this case, $\delta=1$ for illustrative purposes. However, the second derivative (not shown) is discontinuous. It is clear that $\rho_H''(x)$
will be discontinuous at both $-\delta$ and $+\delta$. Hence, the
Hessian is not defined at those points.

\begin{figure}[!ht]
    \centering
    \includegraphics[scale = 0.5]{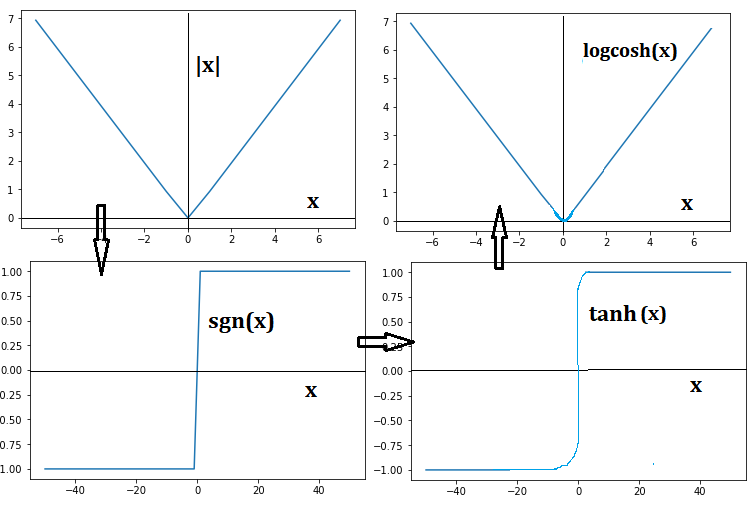}
    \caption{Developing a continuous L1 function.}\label{fig:L1logcosh}
\end{figure}

Another intuitive view of log-cosh can be obtained
by starting with the L1 loss function. 
This loss function produces the median as the estimate but it
has a discontinuous first derivative.
We seek to find a way to construct a continuous function that
has continuous first and second
derivatives to replace the L1 function.
Consider how one might do this intuitively starting with the L1 loss function. 
This is shown in Fig. \ref{fig:L1logcosh}. The absolute value function, $|x|$, has a ``V'' shape. When
we take its derivative, we obtain sgn($x$) which is discontinuous at $x=0$.
We can replace the discontinuous derivative function
with a continuous version using $\tanh(x)$. 
They look about the same but we know that 
one is discontinuous and the other is continuous.
Then, by integrating the $\tanh(x)$ function, 
we obtain $\log(\cosh(x))$ which
does not have a kink (although it may appear to have one in the figure).

\section{Statistical Properties}
Consider the general form of the Cosh distribution of Eqn. \eqref{eqn1}.
The MLE of $\theta$ can be derived as given in Eqn. \eqref{eqn060}. 
The estimator is represented as $\hat\theta$. Then, as $n \rightarrow \infty$,
we find that
\begin{equation}\label{eqn7}
\sqrt{n} (\hat\theta-\theta) \overset{\mathcal{D}}{\to} \mathcal{N}(0,\mathcal{I}^{-1}(\theta))
\end{equation}
where $\mathcal{I}(\theta) $ is the Fisher information which
can be derived as
\begin{equation}\label{eqn14}
\mathcal{I}(\theta) = \frac1{2\sigma^2}.
\end{equation}
Furthermore, the MLE estimator is consistent such that
\begin{equation}\label{eqn17}
\hat\theta \overset{\mathcal{P}}{\to} \theta
\end{equation}
with variance
$$\text{Var}(\hat\theta)=\frac{1}{n \mathcal{I}(\theta)}.$$
Hence, we find the asymptotic variance of $\hat\theta$ to be
\begin{equation}\label{eqn18}
\text{Var}(\hat\theta) =\frac{2\sigma^2}{n}.
\end{equation} 
Next, given that $\hat\theta \overset{\mathcal{P}}{\to}\theta$, the asymptotic bias is
$$\text{bias} = \mathbb{E}[\hat\theta]-\theta=0.$$ The log-cosh estimator
is therefore asymptotically unbiased. Further, the $(1-\alpha)$ confidence interval is given by
$$CI = \bigg[\hat\theta -  \frac{z_{\alpha/2}}{\sqrt{n\mathcal{I}(\theta)}},\,\, \hat\theta +  \frac{z_{\alpha/2}}{\sqrt{n \mathcal{I}(\theta)}}\bigg].$$

\begin{figure}[!ht]
    \centering
    \includegraphics[scale = 0.5]{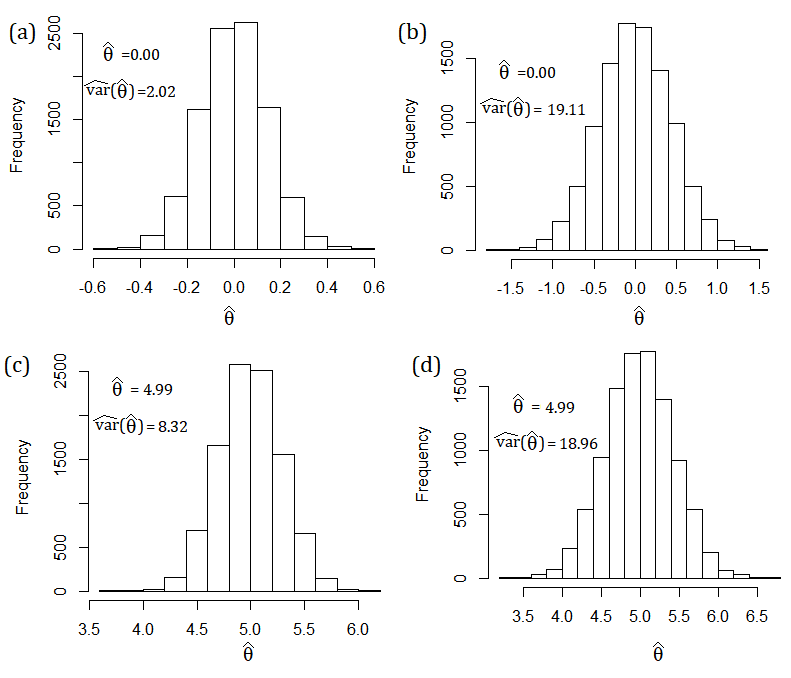}
    \caption{Histogram of estimates of $\hat\theta$ and $n\widehat{\text{Var}}(\hat\theta)$ from 10000 samples  with $n=100$.}\label{fig:MLEdist}
\end{figure}

We can use the bootstrap method to validate the results. Specifically, we
select a sample of size $n$ from a uniform distribution
$\mathcal{U}[0,1]$, say $(u_1, u_2, ..., u_n)$, to produce the set
$x_i = F^{-1}(u_i)$, where $F(x)$ is defined in Eqn. \eqref{eqn01}. 
Its inverse is given by
\begin{equation}\label{eqn010}
F^{-1}(u) =  \sigma\,\sinh^{-1}(\text{tan}\big[(u-\frac12)\pi\big])+\theta.
\end{equation}
The bootstrapped distributions generated in this manner are shown
in Fig. \ref{fig:MLEdist}.
A number of different combinations of $\theta$ and $\sigma$ were selected.
For each case, estimates of the location parameter, $\hat\theta$, and
the variance term, $n\widehat{\text{Var}}(\hat\theta)$, are obtained
and listed in the figure. The same results are
also provided in Table \ref{tab:bootvars} along with the 
estimate of $\hat\sigma$. The estimates in all cases validate the equations
derived earlier (see Eqn. \eqref{eqn18}).
\begin{table}[htbp]
\caption{Locationand Variance estimates using bootstrapping.}
\begin{center}\label{tab:bootvars}
\begin{tabular}{|c|c|c|c|c|c|}
\hline
 &\multicolumn{5}{|c|}{\textbf{Location/Variance}} \\
\cline{2-6} 
 \textbf{Plot}& & & & & \\
 label& $\theta$ & $\sigma$ & $\hat\theta$ & $n\widehat{\text{Var}}(\hat\theta)$ & $\hat\sigma$\\
\hline
& & & & & \\
(a)& 0.0 & 1.0 &  0.00 & 2.02 & 1.00 \\
\hline
& & & & &  \\
(b)& 0.0 & 3.0 &  0.00 & 19.11 & 3.09\\
\hline
& & & & & \\
(c)& 5.0 & 2.0 & 4.99 & 8.32 & 2.04\\
\hline
& & & & & \\
(d)&  5.0 & 3.0 & 4.99 & 18.96 & 3.08\\
\hline
\end{tabular}
\label{tab2}
\end{center}
\end{table}
\section{Comparison with Other Distributions}
\begin{figure}[!ht]
    \centering
    \includegraphics[scale = 0.5]{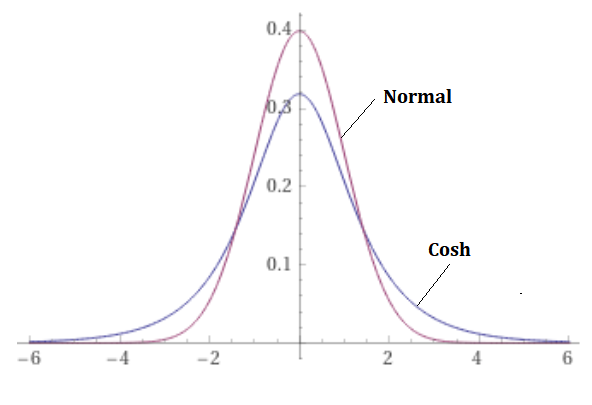}
    \caption{Comparison of standard Normal distribution with Cosh distribution.}\label{fig:NormalCosh}
\end{figure}

\begin{figure}[!ht]
    \centering
    \includegraphics[scale = 0.9]{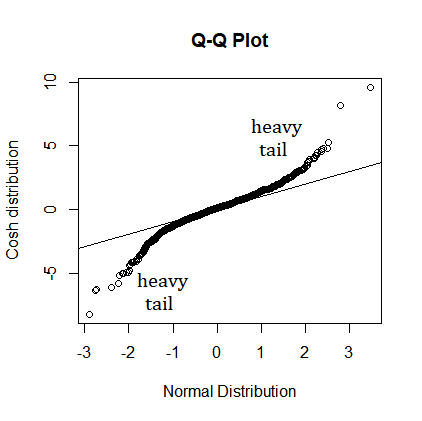}
    \caption{Q-Q plot of Cosh against Normal distribution.}\label{fig:qqplot}
\end{figure}

Our next step is to compare the Cosh distribution with the Normal and
Cauchy distributions. We begin with a comparison with the standard Normal case.
We see in Fig. \ref{fig:NormalCosh} that the Cosh distribution has
heavier tails than the Gaussian $\mathcal{N}(0,1)$ distribution.
This is also illustrated in the Q-Q plot of Fig. \ref{fig:qqplot}
for a random set of points from the Cosh distribution.

\begin{figure}[!ht]
    \centering
    \includegraphics[scale = 0.5]{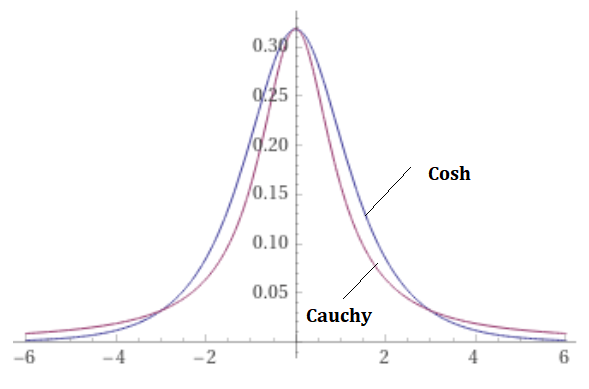}
    \caption{Comparison of Cauchy distribution with Cosh distribution.}\label{fig:CauchyCosh}
\end{figure}

The second comparison is with the Cauchy distribution. One may recognize the form of the pdf of the Cosh distribution as being
similar to the Cauchy distribution (Student t distribution with df=1) given by:
\begin{equation}\label{eqn22}
f(x;\theta,\sigma) =  \frac{1}{\pi\sigma\,(1+(\frac{x-\theta)}{\sigma})^2)}
\end{equation}
with corresponding cdf given by
\begin{equation}\label{eqn23}
F(x;\theta,\sigma) =  \frac12+ \frac1\pi{\tan^{-1}\bigg(\frac{x-\theta}{\sigma}\bigg)}.
\end{equation}
We note here that for the Cauchy distribution
$$\mathbb{E}[X] = \text{undefined}$$
and
$$\text{Var}(X) = \text{undefined}.$$
This may be one of the reasons why the Cauchy distribution is not used
in practice.
A graphical comparison of the Cosh and Cauchy distributions is provided
in Fig. \ref{fig:CauchyCosh}.

We can still develop an estimator based on the Cauchy distribution.
Applying MLE produces the following estimator:
\begin{equation}\label{eqn066}
\hat \theta =  \text{argmin}_{\theta \in \mathbb{R}} \sum_{i=1}^n\text{log} (1+(x_i-\theta)^2)
\end{equation}
The associated loss function can be shown to be non-convex so it is not 
suitable for regression. However, there are simple cases for which it
will produce an acceptable result.
\begin{figure}[!ht]
    \centering
    \includegraphics[scale = 0.65]{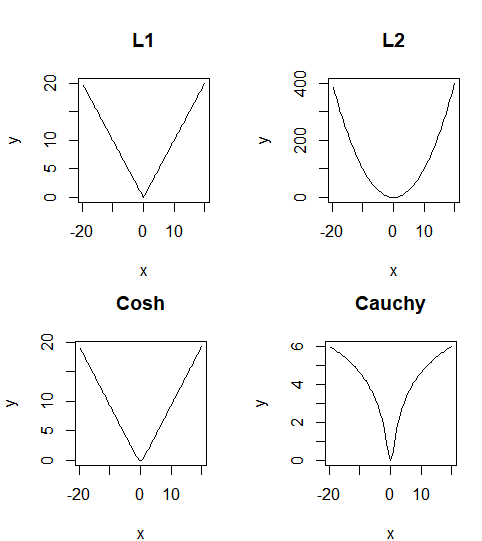}
    \caption{$\rho(x)$ function.}\label{fig:rho}
\end{figure}
\begin{figure}[!ht]
    \centering
    \includegraphics[scale = 0.65]{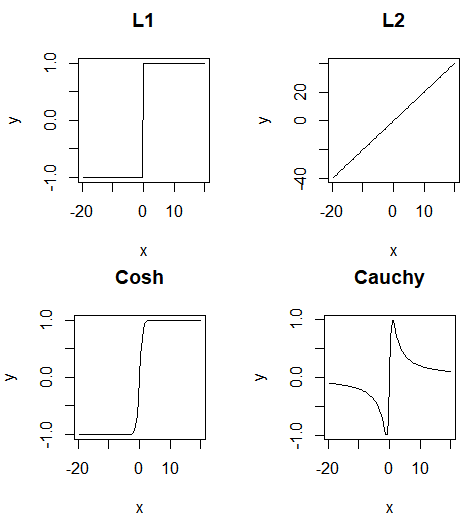}
    \caption{$\psi(x)$ function.}\label{fig:psi}
\end{figure}

Figs. \ref{fig:rho} and \ref{fig:psi} show the characteristics of the $\rho(x)$ and $\psi(x)$ functions derived from the double exponential (L1), 
Gaussian (L2), Cosh and Cauchy distributions. Comparing the different graphs, it
is clear that the Cosh characteristics are very similar to the L1 characteristics.
It inherits some of the good properties of L1 without the discontinuities 
in the first derivative. This is the main reason for its recent rise in
popularity in the machine learning community.
\section{Regression Examples}
\subsection{Location Problem}
Consider a location problem of the form:
\begin{equation}
y_i = \theta + \epsilon_i, \quad i = 1,\ldots,n.  \nonumber
\end{equation}
The one-dimensional data set under consideration, ranging from -2.8 to 20.5, is given in 
Table \ref{tab:ch1:3:dataset}. For this set,
the LS, Cauchy and log-cosh estimates for $\theta$ were computed.

\begin{figure}[!ht]
    \centering
    \includegraphics[scale = 0.4]{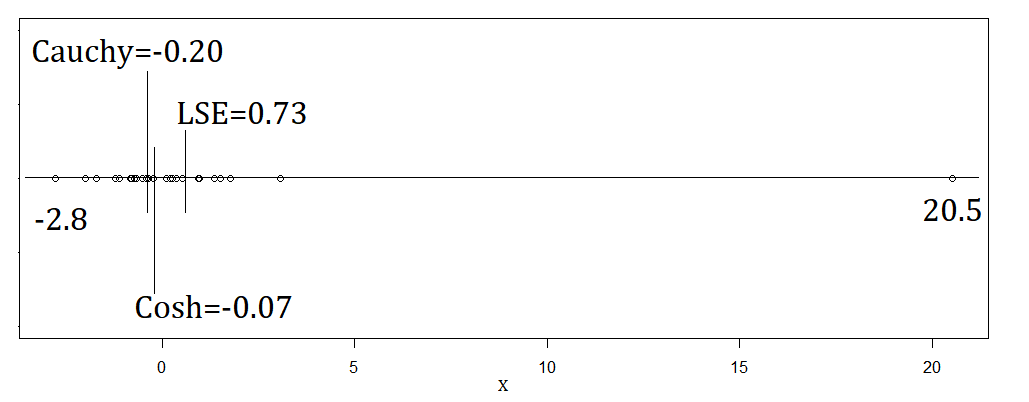}
    \caption{Estimates for Cauchy, Cosh and LSE on location problem.}\label{fig:location}
\end{figure}

\begin{table}[!htbp]
\begin{center}
\caption{Location problem data set.} \label{tab:ch1:3:dataset}
\resizebox{8cm}{!} {
\begin{tabular}{ r | l | l | l | l | l | l }
\hline
(1) & -2.80 & -1.98 & -1.70 & -1.20 & -1.10 & -0.82 \\
\hline
(7) & -0.79 & -0.73 & -0.66 & -0.51 & -0.41 & -0.35 \\
\hline
(13) &  -0.23 &  0.10 &  0.22 &  0.25 &  0.37 &  0.52\\
\hline
(19) &  0.93 & 0.95 & 1.36 & 1.52 & 1.76 & 3.07 \\
\hline
(25) &  20.50 & - & - & - & - & -\\

\hline
\end{tabular}
}
\end{center}
\end{table}
The results are provided in Table \ref{tab:location} and illustrated
graphically in Fig. \ref{fig:location}. We see that the estimates
of Cauchy and log-cosh are quite close in value whereas the LSE is 
much further away due to the outlier on the right-hand side.
Asymptotically, the Cauchy estimator should produce the median, which in
this case is -0.23 while the mean is 0.73. 
We also note the standard errors (s.e.'s), obtained
using bootstrapping and given
in Table \ref{tab:location}, indicate a 3 times larger value for LSE
compared to the relatively smaller values for Cauchy and log-cosh.
\begin{table}[htbp]
\caption{Comparison of LSE, Cosh and Cauchy for location problem $n=25$.}
\begin{center}\label{tab:location}
\begin{tabular}{|c|c|c|c|}
\hline
\textbf{Statistic}&\multicolumn{3}{|c|}{\textbf{Technique Used}} \\
\cline{2-4} 
\textbf{Used} & \textbf{\textit{LSE}}& \textbf{\textit{Cosh}}& \textbf{\textit{Cauchy}} \\
\hline
$\hat\theta$& 0.73 & -0.06 &  -0.19 \\
\hline
s.e.($\hat\theta$)& 0.84 & 0.28 &  0.28 \\
\hline
\end{tabular}
\label{tab2}
\end{center}
\end{table}
\subsection{Simple Linear Regression Example}
We will now compare a number of different methods on a simple linear example
from the \textit{telephone} data set in the \texttt{Rfit} package \cite{b22} in R. 
Rfit uses another robust method called rank-based regression \cite{b9} \cite{b25} to be
described shortly.
For our purposes, we will use the Belgium telephone data set \cite{b22}\cite{b9}. This data set is provided in Table \ref{tab:ch1:2:dataset} and contains the number of calls (in units of 10's of millions) made in Belgium in the years between 1950 - 1973. It has $24$ data points, with $6$ outliers. The outliers are due to a change in measurement technique without re-calibration for $6$ years, as is often cited.

\begin{table}[!htbp]
\begin{center}
\caption{Belgium telephone data set.} \label{tab:ch1:2:dataset}
\resizebox{8.5cm}{!} {
\begin{tabular}{ l|l | l | l | l | l | l | l | l | l | l | l | l}
\hline
x & 1950 & 1951 & 1952 & 1953 & 1954 & 1955 & 1956 & 1957 \\
\hline
y & 0.44 & 0.47 & 0.47 & 0.59 & 0.66 & 9.73 & 0.81 & 0.88 \\
\hline
x &  1958 & 1959 & 1960 & 1961 & 1962 & 1963 & 1964 & 1965\\
\hline
y &  1.06 & 1.2 & 1.35 & 1.49 & 1.61 & 2.12 & 11.9 & 12.4\\
\hline
x  & 1966 & 1967 & 1968 & 1969 & 1970 & 1971 & 1972 & 1973 \\
\hline
y & 14.2 & 15.9 & 18.2 & 21.2 & 4.3 & 2.4 & 2.7 & 2.9 \\
\hline
\end{tabular}
}
\end{center}
\end{table}
Rank-based methods \cite{b9}\cite{b25} also offer robust regression so it is useful
to compare log-cosh against this approach. The rank-based loss function is
governed by:
\begin{equation}
\text{loss}_R(\beta) = \sum_{i = 1}^n (y_i - x_i\beta) a_{n}(R_{n_i}(\beta))
\end{equation}
where $R_{n_i}(\beta)$ are the ranks of the residuals $(y_i - x_i\beta)$ and
\begin{equation}
    a_n(R_{n_i}(\beta)) = \phi\left( \frac{R_{n_i}(\beta)}{n+1} \right)
\end{equation}
with $\phi(u) = 2u-1$. Further details of these quantities may
be found in \cite{b9}\cite{b25}. 

\begin{figure}[!ht]
    \centering
    \includegraphics[scale = 1.0]{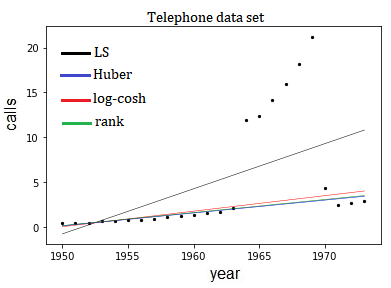}
    \caption{Results for LSE, log-cosh, rank and Huber on a simple linear problem.}\label{fig:simple}
\end{figure}

The results of the regression for LS, Huber, log-cosh, and rank are shown in
Fig. \ref{fig:simple}. The estimates are given in 
Table \ref{tab:estimates}. We note that log-cosh, Huber and rank  are all aligned whereas
the LSE model is greatly affected by the outliers. This example clearly
illustrates the robustness property of the log-cosh estimator. In addition,
the Huber and rank results are very similar in this case, while the
log-cosh method produces a slightly higher slope. This will vary from dataset to dataset.

\begin{table}[!htbp]
\begin{center}
\caption{Slope and intercept for 4 regression methods on telephone data set.} \label{tab:estimates}
\resizebox{6cm}{!} {
\begin{tabular}{ l|l |l}
\hline
 & $\beta_1$  & $\beta_0$ \\
\hline
least squares & 0.504 & -983.9  \\
\hline
log-cosh &  0.173 & -338.1 \\
\hline
rank-based  & 0.146 & -284.3   \\
\hline
Huber ($\delta=0.1$) &  0.143 & -280.0  \\
\hline
\end{tabular}
}
\end{center}
\end{table}
\begin{figure}[!ht]
    \centering
    \includegraphics[scale = 0.85]{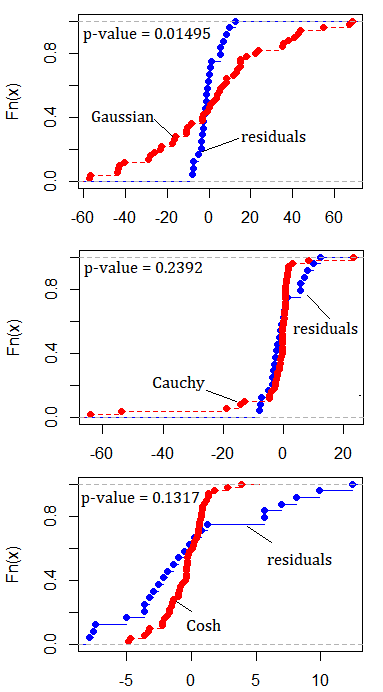}
    \caption{Komlogorov-Smirnov Goodness-of-Fit Test using residuals of Telephone data set comparing p-values of Gaussian, Cauchy and Cosh distributions.}\label{fig:KStest}
\end{figure}
A Goodness-of-fit test can be performed using the Komlogorov-Smirov (K-S) method to examine what distribution best fits the residuals of the telephone 
data set following least squares estimation (LSE). This test uses
the empirical cdf of the distributions. The results are shown graphically
in Fig. \ref{fig:KStest} along with their p-values. We note that the null hypothesis is rejected in the case of the Gaussian distribution at the $\alpha=0.05$ 
level whereas it cannot be rejected for the Cauchy or Cosh distributions.
Hence, one should use a robust method. If the
null hypothesis fails, one should consider a robust method. On the other hand,
the log-cosh loss provides equivalent estimates in cases when outliers
are not present in the data set. Therefore, it may
be used whether or not outliers exist.
\section{Multiple Linear Regression}
The different approaches for robust regression can be compared in
the context of multiple linear regression to study standard errors of the estimates.
We use a well-known Swiss Fertility data set with 5 variables and
47 observations. This data set is part of the R environment. 
The data itself may also be found in \cite{b9}. The variable abbreviations are
as follows: A = Agriculture, Ex = Examination, Ed = Education, C = Catholic, IM = Infant Mortality. The goal is to build a model that predicts Fertility
based on these five explanatory variables. In Table \ref{tab8}, we provide the
estimates for Huber, Rank and log-cosh, along with their associated
standard errors (s.e.). They all produce similar s.e. values, although
the results may vary slightly from dataset to dataset.
\begin{table}[htbp]
\caption{Estimates and standard errors $(s.e.)$ for Huber (H), Rank (R) and log-cosh (L) on Swiss data set.}
\begin{center}\label{tab:vars}
\begin{tabular}{|c|c|c|c|c|c|c|}
\hline
 & & & & & & \\
 & $\hat\beta_n^H$ & s.e.($\hat\beta_n^H$) &$\hat\beta_n^R$ & s.e.($\hat\beta_n^R$)& $\hat\beta_n^L$ & s.e.($\hat\beta_n^L$) \\
  & & & & & & \\
\hline
A   &   -0.19  & 0.071 &  -0.20 &  0.069   & -0.20 &  0.075 \\
Ex  &    -0.28 &  0.258  &  -0.25 &  0.249 & -0.26  & 0.264\\
Ed  &     -0.84 &  0.186 &   -0.88 &  0.179 & -0.89 &  0.190\\
C    &      0.10 &  0.035  &  0.10  & 0.034 & 0.10  & 0.035\\
IM  &       1.21 & 0.388  &  1.19 &  0.375 & 1.40  & 0.395 \\
\hline
\end{tabular}
\label{tab8}
\end{center}
\end{table}
\section{Quantile Regression}
Quantile regression is a robust method for studying the effect of
explanatory variables on the entire
conditional distribution of the response variable
rather than just on the median.
It has been used extensively since the initial concepts
were developed in the late 1970's and early 1980's \cite{b10}. It is based on the so-called check function
whereby the quantile of interest is
set by a parameter, $\tau$. The original check function
exhibits a non-monotonic behavior which has been the subject of
a number of research papers over the years \cite{b30}\cite{b31}\cite{b32}\cite{b36}\cite{b37}. 
In \cite{b5}, we used the log-cosh function to develop a new M-estimator to overcome the crossing problem.
In this section, we derive an MLE equivalent so that a statistical
analysis can be carried out.

\begin{figure}[!ht]
    \centering
    \includegraphics[scale = 0.5]{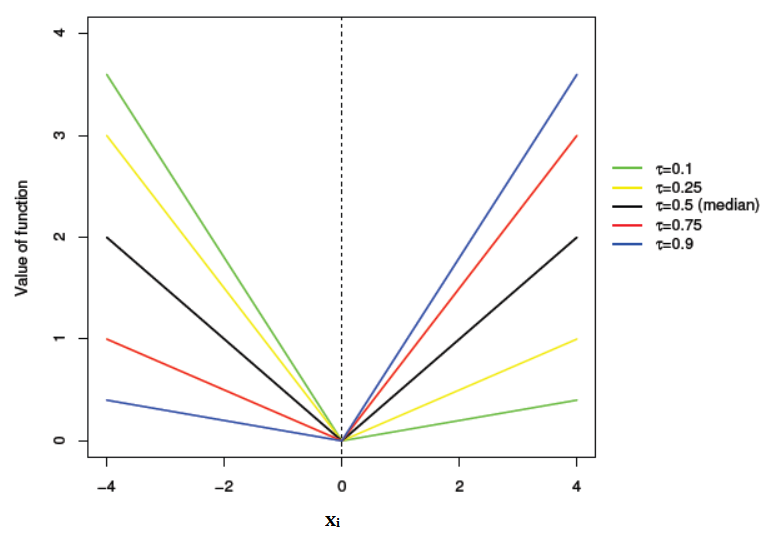}
    \caption{Check function. }\label{fig:check1}
\end{figure}

The original check function can be written concisely as follows:
\begin{equation}\label{eq:check1}
\rho_\tau(x_i,\tau) =
  \begin{cases}
    -(1-\tau) x_i       & \quad \text{if } x_i < 0\\
    \tau x_i  & \quad \text{if } x_i \ge 0
  \end{cases}
\end{equation}
where the parameter $ \tau \in (0,1)$ is the quantile 
and $x_i$ is the $i$th residual.
Different regression quantiles represented by 
lines or hyperplanes are obtained by selecting $\tau$ and 
minimizing the conditional quantile function
\begin{equation}
Q_\tau(Y|x,\tau)=\sum_{i=1}^n \rho_\tau(x_i,\tau).
\end{equation} 
A plot of $\rho_\tau(x_i,\tau)$ for different values of $\tau$ is provided in Fig. \ref{fig:check1}.
Suppose we are interested in the median, which is also the 50th percentile and the
2nd quartile. 
We would set $\tau=0.5$ and use the following check function,
\begin{equation}\label{eq:check11}
\rho_\tau(x_i,0.5) =
  \begin{cases}
    -(0.5) x_i       & \quad \text{if } x_i < 0\\
    0.5 x_i  & \quad \text{if } x_i \ge 0.
  \end{cases}
\end{equation}
This is equivalent to the least absolute deviation (LAD) function or the L1 loss
function.
We see from the figure (black line, $\tau=0.5$) that it is symmetric about
$x_i=0$ but
has a kink at 0. This is the same characteristic that causes problems
for the L1 function. Because of this kink, the derivative of this function
is discontinuous at 0. Other cases shown in the figure for 
$\tau=0.1 ,0.25,0.75,0.9$ all exhibit a kink except that the 
function is asymmetric in one direction or the other. The red lines associated
with $\tau=0.75$ are reminiscent of a check mark, and hence the name
\textit{check function}. In any case, the kinks in this function imply discontinuous derivatives leading to
a number of mathematical and numerical problems when solving
for quantiles.

\subsection{Continuous Check Function}
To circumvent this problem, one could employ a continuous check function 
using log-cosh as follows, 
\begin{equation}\label{eqn50}
\rho_S(x,\tau) =  \text{log}(\text{cosh}( x))+(\tau-\frac12)x
\end{equation} 
This function does not have any kinks as will be demonstrated
shortly, but we can already anticipate that it will be smoother
simply due to the characteristics of log-cosh.

The quantile regression problem involves minimizing an associated 
convex loss function which is the conditional quantile function
for each $\tau$ given by
\begin{equation}\label{eqn500}
Q_S(Y|x,\tau)=\sum_{i=1}^n \rho_S(r_i,\tau).
\end{equation} 

To derive this continuous check function,
we first postulate a pdf given by
\begin{equation}\label{eqn51}
f(x) =  \frac{e^{-(\tau-\frac12)x}}{\kappa\,\text{cosh}(x)}
\end{equation}
where $\kappa$ is a normalizing constant term such that the 
pdf integrates to 1. The value of $\kappa$
will vary as a function of the selected $\tau$.
A table of values for $\kappa$ for selected $\tau$ values is given
in Table \ref{tab:kappa}. Note that for $\tau=0.5$, we obtain
that $\kappa=\pi$ as expected, since it is the original log-cosh
pdf.
\begin{table}[!htbp]
\begin{center}
\caption{Selected values of $\kappa$} \label{tab:kappa}
\resizebox{3cm}{!} {
\begin{tabular}{ r|l }
\hline
$\tau$ & $\kappa$  \\
\hline
0.0 & $\pi\sqrt{2}$  \\
\hline
0.25 &  $\pi\sqrt{4-2\sqrt{2}}$ \\
\hline
0.5 &  $\pi$ \\
\hline
0.75  & $\pi\sqrt{4-2\sqrt{2}}$  \\
\hline
1.0 & $\pi\sqrt{2}$ \\
\hline
\end{tabular}
}
\end{center}
\end{table}

The MLE for this distribution can be derived as follows:
\begin{equation}\label{eqn52}
L(x_1,...,x_n) =  \bigg(\frac{e^{-(\tau-\frac12)x_1}}{\kappa\,\text{cosh}(x_1)}\bigg)\dots\bigg(\frac{e^{-(\tau-\frac12)x_n}}{\kappa\,\text{cosh}(x_n)}\bigg).
\end{equation}
Then,
\begin{equation}\label{eqn53}
-\ell(x_1,...,x_n) =  \sum_{i=1}^n[\text{log(cosh}(x_i))+(\tau-\frac12)x_i + \text{log}(\kappa)]
\end{equation}
Therefore, after removing the constant term, we obtain Eqns. \eqref{eqn50} and \eqref{eqn500}.
The Fisher information can be derived to be
\begin{equation}\label{eqn54}
\mathcal{I}(\theta) = \frac1{2}.
\end{equation}
\begin{figure}[!ht]
    \centering
    \includegraphics[scale = 0.5]{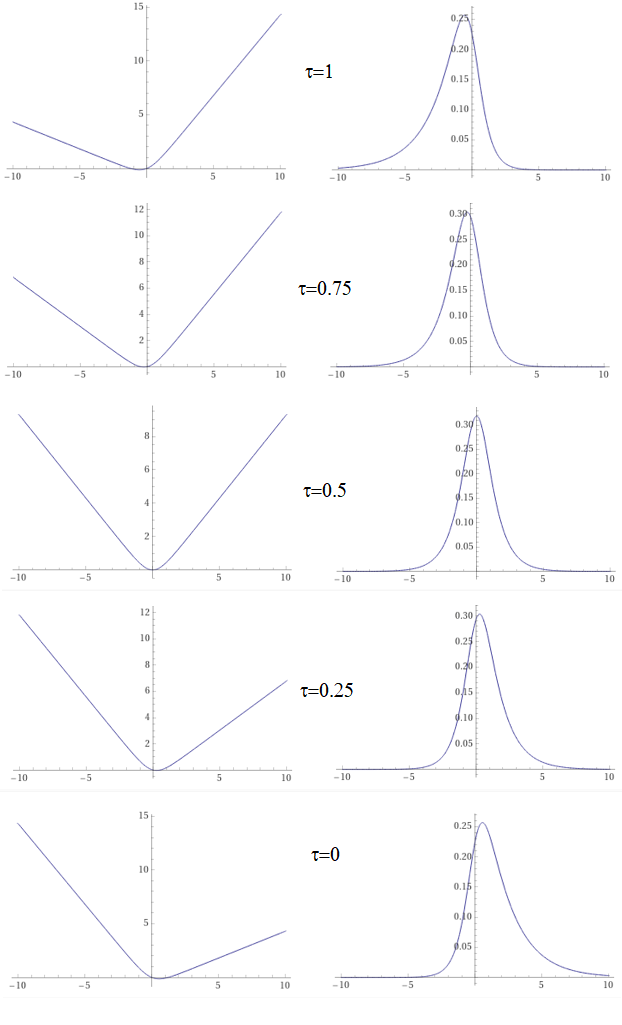}
    \caption{pdf of quantile function.}\label{fig:quantpdf}
\end{figure}

The distributions and loss functions can now be plotted.
Starting with Eqn. \eqref{eqn51} and the normalizing constants $\kappa$ given in Table \ref{tab:kappa}, different values of $\tau$ were selected and their associated distributions plotted in Fig. \ref{fig:quantpdf}.
The cases shown are for $\tau=0, 0.25, 0.5, 0.75, 1.0$. The distributions
shown on the right-hand side of the figure can be seen to skew from
one side to the other as $\tau$ decreases in value from 1 to 0.
On the left-hand side, we have plotted the loss term of the MLE in each case.
We see that the use of the logcosh function smooths out the kinks of the original loss function
(see Fig. \ref{fig:check1}).

A more general
M-estimator based on log-cosh for quantile regression is given by: 
\begin{equation}\label{eqn20}
\rho_{SMRQ}(x,\tau) =  \frac{1}{2c}\text{log}(\text{cosh}(c (x-h)))+(\tau-s)x+v
\end{equation} 
where $c$ is the desired curvature of the function
(i.e. the severity of the kink), $v$ controls vertical shift, $h$ controls the
horizontal shift and $\tau-s$ is
used to produce the asymptotic
slopes on the two sides of the check function itself.

\begin{figure}[!ht]
    \centering
    \includegraphics[scale = 0.53]{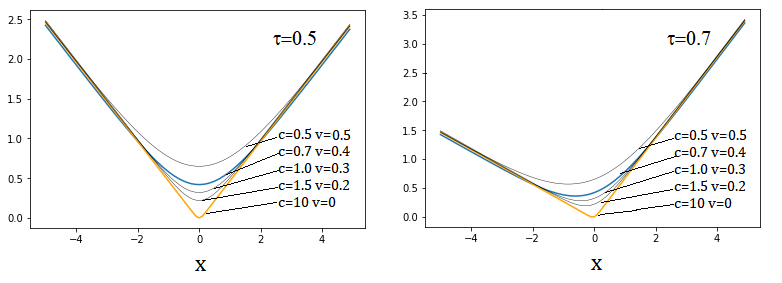}
    \caption{Loss functions for SMRQ for $\tau=0.5$ and $\tau=0.7$ cases.}\label{fig:logcoshnew}
\end{figure}

Quantile regression using the above equation
is referred to as SMRQ for `smoother regression quantiles'.
Fig. \ref{fig:logcoshnew} shows five different cases of Eqn. \eqref{eqn20} by varying 
parameters $c$ and $v$ (while holding $s=0.5$ and $h=0$ fixed) for
$\tau=0.5$ and $\tau=0.7$. One can observe the different levels of
smoothness offered by the flexible check function, which can be varied easily
as the need arises.

We propose the use of $c=1/2$ and $v=1/2$ to produce:
\begin{equation}\label{eqn21}
\rho_{SMRQ}(x,\tau) =  \text{log}(\text{cosh}\big(\frac{x}{2}\big))+(\tau-\frac{1}{2})x+\frac12.
\end{equation}

The variance of the estimates due to this loss function cannot be obtained
in closed form. Therefore, a bootstrapping method can be
used to extract standard errors for estimates, which are naturally
related to the variance.

\subsection{Comparison with Convolutional Smoothing}
Recently, a technique was published \cite{b14} that utilizes convolutional
smoothing \cite{b23} applied to the loss function and is implemented in the \texttt{conquer} library in R. 
Convolutional smoothing provides a different approach to reducing the
crossing problem. Therefore, it is appropriate to compare it with SMRQ 
in terms of solving the crossing problem.

One natural way to monitor the monotonicity of a quantile regression
method is to simply count the number of data points below a given regression
line (or hyperplane in the multivariate case). The median line (50th percentile)
should result in half the points below it. The 75th percentile
should results in 75\% of the data below that line, and so on. 
One rule that should be enforced is that a higher percentile will have
more points below it than a lower percentile. That is, we should not
have more points below the 75th percentile than we do at the 80th percentile.
This would violate the basic definition of percentiles, which is not desirable.
However, in the original
formulation of quantile regression \cite{b24}, this was often the case and
was called the \textit{quantile crossing problem}.

\begin{figure}[!ht]
    \centering
    \includegraphics[scale = 0.45]{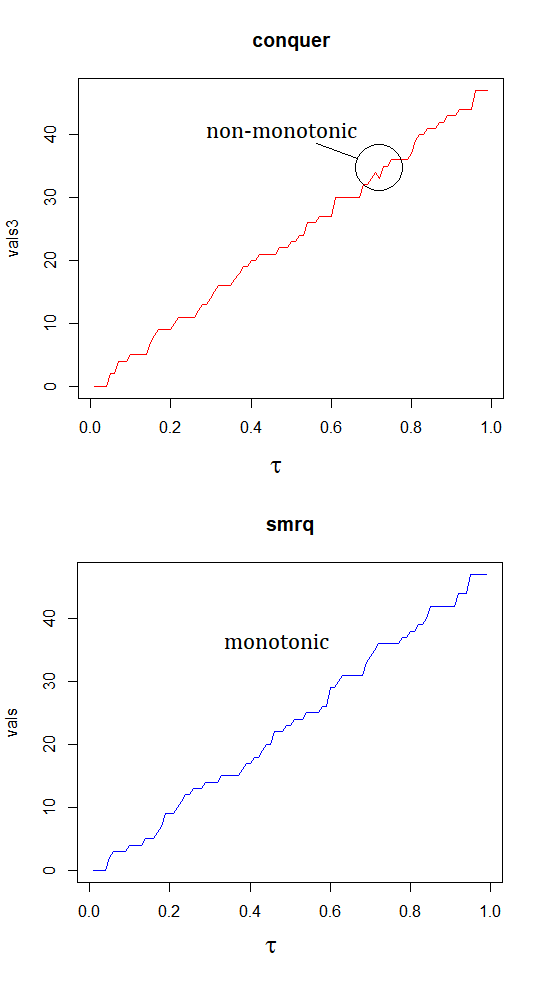}
    \caption{Conquer vs. SMRQ on the swiss data set.}\label{fig:conqvssmrq}
\end{figure}

To investigate the monotonicity of conquer and SMRQ, we selected the well-known
swiss data set in R, which has 5 variables.
We show the  monotonicity results for conquer and SMRQ in 
Fig. \ref{fig:conqvssmrq}.  Note that the plot should be non-decreasing if the
crossing problem is avoided. 
This is true for SMRQ but
conquer exhibits 1 non-monotonic event.

We now compare the standard errors (s.e.) of conquer and SMRQ in Table \ref{tab:conqvars}. Rather than listing each s.e. for all 5 variables
in the 5 cases
of $\tau$ (total of 25 values each), we compute the $L_2$-norm of the s.e.'s and list those values in the table. The numbers 
represent an aggregate standard error for each $\tau$ which is 
sufficient for this comparison.

We see that the errors are on average 10\% smaller for SMRQ compared
to conquer. A similar result was also obtained on the diabetes dataset in R.
In addition,
SMRQ possesses the monotonocity property for quartiles,
deciles and percentiles whereas conquer may exhibit some
non-monotonic behavior.
However, conquer was intended to handle large-scale problems
and requires less runtime for the same size problem.
This is a relatively small example for conquer, but
it is expected to perform much better with large data sets.

\begin{table}[htbp]
\caption{s.e. for conquest and SMRQ as a function of $\tau$ on swiss data set.}
\begin{center}\label{tab:conqvars}
\begin{tabular}{|c|c|c|}
\hline

 \textbf{Quantile}& conquer (C) & SMRQ (S)\\
 \hline
$\tau$ & $|| (s.e._C)||_2$ & $||(s.e._S)||_2$  \\

\hline
& & \\
0.01& 0.851 & 0.781  \\
\hline
& & \\
0.25 & 0.781 & 0.792 \\
\hline
& & \\
0.5 & 0.755 & 0.710 \\
\hline
& & \\
0.75&  0.921 & 0.756  \\
\hline
& & \\
0.99&  0.825 &  0.753  \\
\hline
\end{tabular}
\label{tab2}
\end{center}
\end{table}
\section{Conclusions}
In this paper, we identified the Cosh distribution
from which the MLE for the log-cosh function can be derived.
We then derived the asymptotic variance, asymptotic bias and
confidence intervals. The log-cosh loss function was compared
to the Huber, rank and LSE loss functions in the case of 
simple linear regression. The estimates and standard errors
were found to be similar for Huber, rank and log-cosh 
for multiple linear regression. Next, the use of log-cosh in
quantile regression was described to resolve the 
crossing problem, the details of which can be found in
\cite{b5}. The M-estimator for quantile regression was compared and
found to have a smaller standard error than conquer, which uses convolutional smoothing. 
From the analysis provided herein, it is clear that
the log-cosh is an important loss function for machine learning
and is expected to increase in use over the coming years.

\end{document}